\documentclass[runningheads]{llncs}

\usepackage{graphicx}
\usepackage{booktabs}
\usepackage{color}
\usepackage{booktabs}
\usepackage{makecell}

\usepackage{bbding}
\usepackage{overpic}
\usepackage{bm}
\usepackage{amsmath,array,arydshln}
\usepackage{multicol}
\usepackage{multirow}
\usepackage{hyperref}
\usepackage{amsfonts}
\usepackage{cite}
\usepackage{subfigure}
\usepackage[T1]{fontenc}

\newcommand{\ve}[1]{\mathbf{#1}}
\newcommand{\vv}[1]{\mbox{\boldmath $#1$}}

\begin{document}
\title{Task-oriented Self-supervised Learning for Anomaly Detection in Electroencephalography}
\titlerunning{Task-oriented SSL for Anomaly Detection in EEG}
\author{Yaojia Zheng\inst{1,2} \and
Zhouwu Liu\inst{1,2} \and
Rong Mo\inst{3} \and
Ziyi Chen\inst{3} \and
Wei-shi Zheng\inst{1,2} \and
Ruixuan Wang\inst{1,2}\textsuperscript{(\Envelope)}}

\authorrunning{Y.Zheng et al.}

\institute{School of Computer Science and Engineering, Sun Yat-sen University, China \and
Key Laboratory of Machine Intelligence and Advanced Computing, MOE,  China
\email{wangruix5@mail.sysu.edu.cn} \and
The First Affiliated Hospital, Sun Yat-sen University, Guangzhou, China}
\maketitle              % typeset the header of the contribution

\begin{abstract}

%Electroencephalography (EEG) has been widely used to monitor and diagnose patients with various brain diseases. 
Accurate automated analysis of electroencephalography (EEG) would largely help clinicians effectively monitor and diagnose patients with various brain diseases. Compared to supervised learning with labelled disease EEG data which can train a model to analyze specific diseases but would fail to monitor previously unseen statuses, anomaly detection based on only normal EEGs  can detect any potential anomaly in new EEGs. Different from existing anomaly detection strategies which do not consider any property of unavailable abnormal data during model development, a task-oriented self-supervised learning approach is proposed here which makes use of available normal EEGs and expert knowledge about abnormal EEGs to train a more effective feature extractor for the subsequent development of anomaly detector. In addition, a specific two-branch convolutional neural network with larger kernels is designed as the feature extractor such that it can more easily extract both larger-scale and small-scale features which often appear in unavailable abnormal EEGs. The effectively designed and trained feature extractor has shown to be able to extract better feature representations from EEGs for development of anomaly detector based on normal data and future anomaly detection for new EEGs, as demonstrated on three EEG datasets. The code is available at https://github.com/ironing/EEG-AD.

%simulated abnormal EEGs from normal EEGs
%two-stage framework is proposed here for anomaly detection based on normal EEGs only, where in the first stage a novel  and  
% 

%The abstract should briefly summarize the contents of the paper in 15--250 words.

\keywords{Anomaly detection \and Self-supervised learning \and EEG.}
\end{abstract}

\section{Introduction}
% EEG, roles and applications; 
% related work about anomaly detection in EEGs: mostly supervised, a few based on normal data only; supervised work cannot collect all possible anomalies, this study focus on the latter; 
% existing ad strategies: statistical/generative, discriminative, reconstruction, self-supervised, their combinations;
% all existing strategies did not consider specific properties in abnormal EEGs, this study for the first time propose a two-framework which consider charac of abnormal EEGs during development.
% contributions: framework, sota results

% to compare with CutPaste somewhere

Electroencephalography (EEG) is one type of brain imaging technique and has been widely used to monitor and diagnose brain status of patients with various brain diseases (e.g., epilepsy)~\cite{gemein2020machine,fiest2017prevalence,ItamarMegiddo2016HealthAE,dhar2021brain,alturki2020eeg}.
EEG data typically consists of multiple sequences (or channels) of waveform signals, with each sequence obtained by densely sampling electrical signals of brain activities from a unique electrode attached to a specific position on patient's head surface. While brain activities can be recorded by EEG equipment over hours or even days, clinicians often analyze EEG data at the level of seconds, considering that cycles of most brain activities varies between 0.5Hz and 30Hz. Therefore, it is very tedious to manually analyze long-term EEG data and automated analysis of EEG would largely alleviate clinician efforts in timely monitoring patient statuses. 

Currently, most automated analyses of EEGs focus on specific diseases~\cite{FernandoPrezGarca2021TransferLO,JeffCraley2018ANM,FelixAchilles2018ConvolutionalNN}, where labelled EEGs at the onset of disease and normal (healthy) status are collected to train a classifier for prediction of patient status at the level of seconds. However, such automated systems can only help analyze specific diseases and would fail to recognize novel unhealthy statuses which do not appear during classifier training. %On the other hand, considering that, collecting EEGs of all types of    
In contrast, developing an anomaly detector based on only normal EEGs has the potential to detect any possible unhealthy status (i.e., anomaly) in new EEG data. While multiple anomaly detection strategies have been developed for both natural and medical image analyses~\cite{NinaShvetsova2021AnomalyDI,YuTian2021ConstrainedCD}, 
including statistical approaches~\cite{WatsonJia2019AnomalyDU,OliverRippel2021ModelingTD},
%like Gaussian density distribution, 
discriminative approaches~\cite{RaghavendraChalapathy2018AnomalyDU,wang2019gods}
%like one-class SVM and one-class convolutional neural network~\cite{,}, 
reconstruction approaches~\cite{VitjanZavrtanik2021ReconstructionBI,ThomasSchlegl2019fAnoGANFU}
%like autoencoder and f-AnoGAN~\cite{,}, 
and self-supervised learning approaches~\cite{LiangChen2019SelfsupervisedLF,ZhenyuLi2020SuperpixelMA}, 
%like *** and ***~\cite{,}, 
very limited studies have been investigated on anomaly detection based on normal EEGs only~\cite{JunjieXu2020AnomalyDO}. Furthermore, all these strategies build anomaly detectors without considering any property of anomaly due to absence of abnormal data during model development. One exception is the recently proposed CutPaste method for anomaly detection in natural images~\cite{ChunLiangLi2021CutPasteSL}, where simulated abnormal images were generated by cutting and pasting small image patches in normal images and then used to help train a more effective feature extractor and anomaly detector.
%a deep convolutional feature extractor was first trained based on normal images and simulated abnormal images, and then Gaussian distribution of normal images in the feature space is built to help detect potential anomalies in new images. 

Inspired by the CutPaste method, we propose a novel task-oriented self-supervised learning approach to train an effective feature extractor based on normal EEG data and expert knowledge (key properties including increased amplitude and unusual frequencies) about unavailable abnormal EEGs.
%two-stage framework to develop an anomaly detector based on normal EEG data only. In the first stage, key properties (i.e., increased amplitude and unusual frequencies) of anomaly in EEGs are considered and used to help generate two types of fake abnormal EEG data as part of training set, 
In addition, a specific two-branch convolutional neural network (CNN) with larger kernels is designed for effective extraction of both small-scale and large-scale features, such that the CNN feature extractor can be trained to extract features of both normal and abnormal EEGs. The feature extractor with more powerful  representation ability can help establish a better anomaly detector.
%at the second stage where various anomaly detection strategies can be adopted in practice.
State-of-the-art anomaly detection performance was obtained on one internal and two public  EEG datasets, confirming the effectiveness of the proposed approach. %ablation studies, support that the proposed framework is effective in 

%Extensive empirical evaluations on 

%in different areas outside the brain which appears as waveforms that vary in frequency and amplitude measured in voltages (usually microvoltages). 
%EEG waveforms are often classified according to their frequency, amplitude, and shape. 

\section{Methodology}%Two-stage framework for anomaly detection}
%\subsection{A two-stage framework for Anomaly Detection using self-supervised learning}

% why two-stage? one stage: either generative or discriminative, two stage can combine both; second stage is flexible (although generative is better considering first stage is discriminative), also list difference vs. cutpaste (data type, different ways to generate abnormal data, different model structure)
% general pretraining does not consider specific features of abnormal EEG data
% 

In this study, we try to solve the problem of anomaly detection in EEGs when only normal EEG data is available for training. A two-stage framework is proposed here, where the first stage aims to a train a feature extractor using a novel self-supervised learning method, and the second stage can adopt any existing generative or discriminative method to establish an anomaly detector based on the feature representations from the well-trained feature extractor. 

\subsection{Task-oriented self-supervised learning}
Various self-supervised learning (SSL) strategies have been proposed to train feature extractors for downstream tasks in both natural and medical image analysis. To make a feature extractor more generalizable for downstream tasks, existing SSL strategies are often designed purposely without regard to any specific downstream task. That means, SSL strategies often do not consider characteristics in specific downstream tasks. However, when applying any such SSL technique to an anomaly detection task, the feature extractor based on only normal data would  less likely learn to extract features of abnormal data, %which would limit the representation power of the trained feature extractor 
and therefore may negatively affect model performance in the subsequent anomaly detection task.

\begin{figure}[t] 
\includegraphics[width=\textwidth, height=0.2\textwidth]{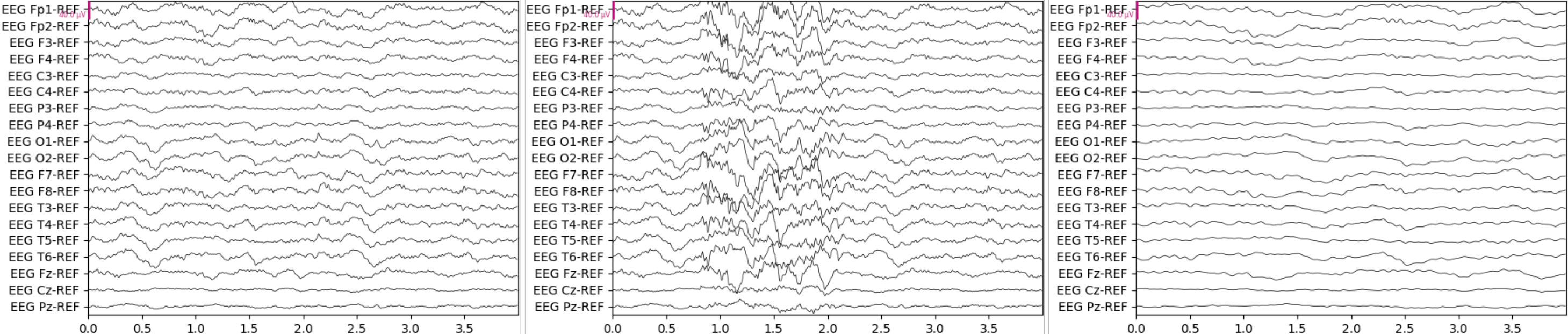}
\caption{ Transformation of normal EEG (Left) to generate simulated amplitude-abnormal (Middle) and frequency-abnormal (Right) EEGs. One EEG data consists of multiple sequences (channels, y-axis) of wave signals. x-axis: time (seconds).
%Left(blue): original normal EEG data; Middle(green): EEG sequences of all channels is multiply with factor $X$; Right(green): All EEG sequences is interpolated by factor $I$ to slow down the frequency.
} 
\label{fig:ab_example}
\end{figure}

Different from most SSL strategies, a novel SSL strategy is proposed here to train a feature extractor which can extract features of both normal and abnormal EEG data. Specifically, considering that abnormal EEGs are characterized by increased wave amplitude or temporally slowed or abrupt wave signals, two special transformations are designed to generate simulated abnormal EEG data. One transformation is to temporally locally increase amplitude of normal EEG signals (Figure~\ref{fig:ab_example}, Middle), and the other transformation is to temporally increase or decrease the frequency of normal EEG signals (Figure~\ref{fig:ab_example}, Right). These amplitude-abnormal and frequency-abnormal data, together with original normal EEGs, form a 3-class dataset for the training of a 3-class CNN classifier. The feature extractor part of the well-trained
classifier would be expected to learn to extract features of both normal and (simulated) abnormal EEGs. While the simulated abnormal data are different from real abnormal EEGs, % and probably cannot represent all kinds of abnormalities in real abnormal EEGs, 
empirical evaluations %(Section~\ref{Effectiveness evaluation}) 
show the feature extractor trained with simulated abnormal EEGs  helps improve the performance of anomaly detection significantly.

% more anomalies in internal set

%  for detection of anomalies in EEG data. Specifically, to train a feature extractor which can extract features of both normal and abnormal EEG data, two types of fake abnoaml EEG data are 

%Following \cite{}, we adopt a two-stage framework, where we firstly learning representation by self-labeled data (self-supervised learning\cite{}) and then utilize Gaussian density estimator\cite{} on deep representations to compute anomaly score.

%We propose a novel self-supervised learning strategy specifically for EEG anomaly detection. Abnormal EEG data, such as sharp and wave complex (see Fig~\ref{fig1}) and spike and wave complex (see Fig~\ref{fig2}), often include wave signals of increasing amplitude and slow wave activities, and seizures (see Fig~\ref{fig3}) are characterized by local or global rhythmic activity with increasing amplitude. Inspired by the observation, we design two different transformations on EEG data: increasing the amplitude of normal EEG waveforms or slow down the frequency and train the classifier to recognize normal and transformed data to encourage the model to learn to detect abnormality.

% \begin{figure}
% \includegraphics[width=\textwidth]{jiman.png}
% \caption{sharp and wave complex.} \label{fig1}
% \end{figure}

% \begin{figure}
% \includegraphics[width=\textwidth]{jianman.png}
% \caption{spike and wave complex.} \label{fig2}
% \end{figure}

% \begin{figure}
% \includegraphics[width=\textwidth]{seizure.png}
% \caption{seizures.} \label{fig3}
% \end{figure}

% \begin{figure}[t] 
% \centering
% \includegraphics[width=0.9\textwidth,height=0.54\textwidth]{multi-abnormal.png}
% \caption{ An example of real abnormal EEGs.} 
% \label{fig:real_EEG}
% \end{figure}

% \begin{figure}[t] 
% \includegraphics[width=\textwidth]{overview-method.png}
% \caption{ An overview of our method for anomaly detection.
% } 
% \label{fig:method}
% \end{figure}

%\vspace{-0.5cm}
\subsubsection{Generation of self-labeled abnormal EEG data:}
\label{SSL-Generation}
% Include an Algorithm table in Supplementary 
Denote a normal EEG data by a matrix $\mathbf{X} \in \mathbb{R}^{K\times L}$, where $K$ represents the number of channels (sequences) and $L$ represents the length of each sequence. To generate an amplitude-abnormal EEG data based on $\mathbf{X}$, a scalar amplitude  factor $\alpha$ is firstly sampled from a predefined range $[\alpha_l, \alpha_h]$ where $1 < \alpha_l < \alpha_h$, followed by sampling a segment length $w$ from a predefined sequence segment range $[w_l, w_h]$ where $1 < w_l < w_h < L$. Then $w$ consecutive columns in $\mathbf{X}$ were randomly chosen, with each element in these columns multiplied by the amplitude factor $\alpha$. Such transformed EEG data with modified $w$ consecutive columns  
%the simple multiplication result $\alpha \mathbf{X}$ 
can be used as a simulated amplitude-abnormal EEG data (Figure~\ref{fig:ab_example}, Middle). On the other hand, to generate a lower-frequency abnormal EEG data based on $\mathbf{X}$, a frequency scalar factor $\omega$ is firstly randomly sampled from a predefined range $[\omega_l, \omega_h]$ where $1 < \omega_l < \omega_h$, and then each sequence (row) of signals in $\mathbf{X}$ is linearly interpolated (i.e., upsampled) by the factor $\omega$ to generate an elongated EEG data $\mathbf{X}' \in \mathbb{R}^{K\times L'}$, where $L' =  \lfloor \omega L \rfloor$ is the largest number which is equal to or smaller than $\omega L$. One frequency-abnormal EEG data can be generated by randomly choosing $L$ consecutive columns from $\mathbf{X}'$. Similarly, to alternatively generate a higher-frequency abnormal EEG data from $\mathbf{X}$, the frequency scalar factor $\omega'$ is firstly randomly sampled from another predefined range $[\omega'_l, \omega'_h]$ where $0 < \omega'_l < \omega'_h < 1$, and then each row of signals in $\mathbf{X}$ is down-sampled by the factor $\omega'$ to generate a shortened EEG data. The shortened EEG data is then concatenated by itself multiple times along the temporal (i.e., row) direction to obtain a temporary EEG data $\mathbf{X}'' \in \mathbb{R}^{K\times L''}$ where $L'' =  \lfloor \omega' L \rfloor \cdot \lceil \frac{1}{\omega'} \rceil \ge L$. One higher-frequency abnormal EEG data can be obtained by randomly choosing $L$ consecutive columns from $\mathbf{X}''$ (when $L'' > L$) or just be $\mathbf{X}''$ (when $L'' = L$). 

Using these simple transformations and based on multiple normal EEG data, two classes of self-labeled abnormal data will be generated, with one class representing anomaly in amplitude, and the other representing anomaly in frequency. %It is worth noting that these simulated abnormal EEGs are generally different from real abnormal EEGs which  are often more irregular and  include combinations of various anomalies. %(Figure~\ref{fig:real_EEG}).
Although more complex transformations can be designed to generate more realistic amplitude- and frequency-abnormal EEGs, empirical evaluations show  the simple transformations are sufficient to help train an anomaly detector for EEGs.

%is represented as a matrix whose dimension is (N × L) where N is the number of channels and L is the sequence length = sampling rate × duration. Channel represent different area of brain. We transform the data sequence of all channels. Our implementation of the augmentation as follows:

%1. Increase the amplitude: We determine the transformed sequence length of the EEG data by sampling from [4, sampling rate * segment duration] and determine the multiplication factor $X$ by sampling from [2.0, 4.0]. Then the location of the beginning of the transformed sequence is randomly selected in a way that the entire sequence appears in the full EEG matrix. Each selected sequence data of all channels will multiply with $X$.

%2. Slow down the frequency: We determine the interpolation factor $I$ by sampling from [2, 4] and then interpolate each sequence of data in the EEG with factor $I$ to produce a longer sequence of $I$ · $d$ values, where $d$ represents the number of values in the original sequence. Then, the $d$ values begin with randomly selected position of the longer sequence are selected to form a new sequence.

%By transform the original normal EEG dataset $\left\{D_{i}\right\}_{i=1}^{N}$, a labeled dataset $\left\{T_{k}\left(D_{i}\right), k\right\}_{i, k=1}^{N, k}$ is generated where $k$ represents the $k$-th transformation ($k$ = 0: original, $k$ = 1: increase the amplitude, $k$ = 2: slow down the frequency). 
%We show the augmentation process in the Fig~\ref{fig4}.

\begin{figure}[t]
\centering
\includegraphics[height=0.17\textwidth]{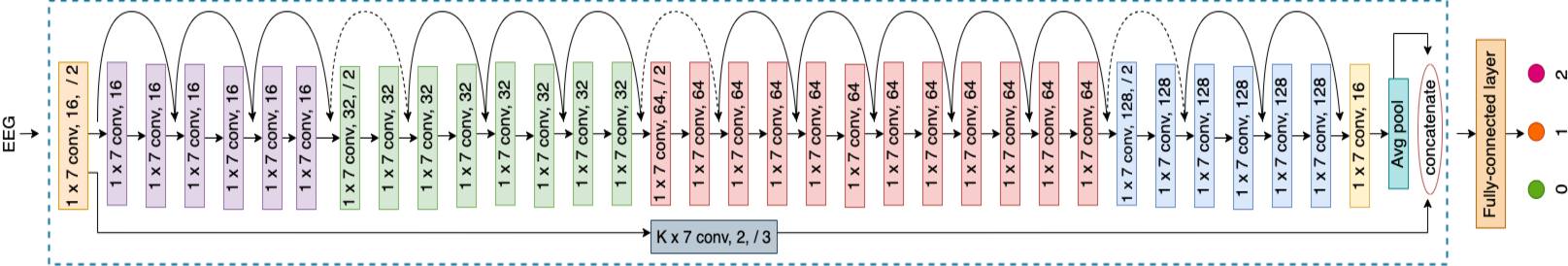}
\caption{The two-branch architecture for feature extractor training. The second branch (lower part) consists of only one convolutional layer for extraction of small-scale features. {The part in the blue dotted box} is the feature extractor.} 
\label{fig:model}
%\vspace{-0.5cm}
\end{figure}

%\vspace{-0.5cm}
\subsubsection{Architecture design for feature extractor:} %Model architecture for prediction of transformations}
To train a feature extractor with more powerful representation ability, we designed a specific CNN classifier based on the ResNet34 backbone (Figure~\ref{fig:model}). Considering that two adjacent channels (rows) in the EEG data do not indicate spatial proximity between two brain regions, one-dimensional (1D) convolutional kernels are adopted to learn to extract features from each channel over convolutional layers as in previous studies~\cite{GuanghaiDai2020HSCNNAC}. However, different from the previously proposed 1D kernels of size $1\times3$, kernels of longer size in time (i.e., $1\times 7$ here) is adopted in this study. Such longer kernels are used considering that lower-frequency features may last for a a longer period (i.e., longer sequence segment) in EEGs and therefore would not be well captured by shorter kernels even over multiple convolutional layers. On the other hand, considering that some other anomalies in EEGs may last for very short time and therefore such abnormal features may be omitted after multiple times of pooling or down-sampling 
%along temporal dimension 
over layers, we propose adding a shortcut branch from the output of the first convolutional layer to the penultimate layer. Specifically, the shortcut branch consists of only one convolutional layer in which each kernel is two-dimensional (i.e., number of EEG channels $\times$ 7) in order to capture potential correlation across all the channels in  short time interval. By combining outputs from the two branches, both small-scale and large-scale features in time would be captured. The concatenated features are fed to the last fully connected layer for prediction of EEG category. The output of the classifier consists of three values, representing the prediction probability of `normal EEG', `amplitude-abnormal EEG', and `frequency-abnormal EEG' class, respectively.   

The feature extractor plus the 3-class classifier head can be trained by minimizing the cross-entropy loss on the 3-class training set. 
After training, the classifier head is removed and the feature extractor is used to extract features from normal EEGs for the development of  anomaly detector. Since the feature extractor is trained without using real abnormal EEGs and the simulated abnormal EEGs are transformed from normal EEGs and self-labelled, the  training is an SSL process. The SSL is task-oriented because it considers the characteristics (i.e., crucial anomaly properties in both amplitude and frequency of abnormal EEGs) in the subsequent specific anomaly detection task. The feature extractor based on such task-oriented SSL is expected to be able to extract features of both normal and abnormal EEGs for more accurate anomaly detection.

%...different from CutPase....

%such details may be useful to distinguish between normal and abnormal signals.

%We trained a deep convolutional neural network (CNN) on the self-labeled dataset $\left\{T_{k}\left(D_{i}\right), k\right\}_{i, k=1}^{N, k}$ to predict the transformation for any augmented EEG data. For EEG data, we propose a novel model architecture which backbone is ResNet34. wave details in very short segment . 

% As shown in Fig~\ref{fig5}, we fuse the high-level features and low-level features because the high-level network has a large receptive field and strong representation ability of semantic information, but it lack of details of spatial geometric features and low-level networks have small receptive fields and strong representation ability of geometric details but the representation ability of semantic information is weak. By this means, we expected it will be helpful for the model to capture coarse- and fine-grained features.

% Furthermore, based on ResNet34 architecture, on account of the long time of slow waves in a period, we replace the size of the convolution kernel with 1 x 7, which is expected to capture features in time dimension of EEG data and helpful to capture as much as possible waveform features within multiple periods.

% After convolving the data with a 1 x 7 filter, we reconvolve the features with a 19 x 7 filter and reduce the output channel to 2, aims to extract feature in time dimension and space dimension and reduce the dimensionality by fusing the feature maps.

% to explain 'task-oriented' ssl

\subsection{Anomaly detection}
While in principle any existing anomaly detection strategy can be applied based on feature representations of normal EEGs from the well-trained feature extractor, here the generative approach is adopted to demonstrate the effectiveness of the proposed two-stage framework considering that a large amount of normal EEG data are available to estimate the distribution of normal EEGs in the feature space. Note discriminative approaches like one-class SVM may be a better choice if normal data is limited. As one simple generative approach, multivariate Gaussian distribution $\mathcal{G}(\vv{\mu}, \ve{\Sigma})$ is used here to represent the distribution of normal EEGs, where the mean $\vv{\mu}$ and the covariance matrix $\ve{\Sigma}$ are directly estimated from the feature vectors of all normal EEGs in the training set, with each vector being the output of the feature extractor given a normal EEG input.

With the Gaussian model $\mathcal{G}(\vv{\mu}, \ve{\Sigma})$, the degree of abnormality for any new EEG data $\ve{z}$ can be estimated based on the Mahalanobis distance between the mean $\vv{\mu}$ and the feature representation $f(\ve{z})$ of the new data $\ve{z}$, i.e.,
\begin{equation}
A(\mathbf{z})=\sqrt{(f(\ve{z})-\vv{\mu})^{\top} \Sigma^{-1}(f(\ve{z})-\vv{\mu})} \,.
\end{equation}
%where $f(\ve{z})$ is the output vector from the feature extractor $f(\cdot)$ for the input $\ve{z}$. 
Larger $A(\mathbf{z})$ score indicates that % the test EEG data 
$\ve{z}$ is more likely abnormal, and vice versa.

%For a testing sample $x$ and a trained classifier $f$, we compute the anomaly score according to the following formula.

%Subsequently, we fit a multivariate Gaussian to deep feature representations of normal data from trained model. The probability density of a test sample which is uniquely determined by Mahalanobis distance becomes the anomaly score.

%After training the model by proxy task, we believe that the model can better learn the boundary between normal data and transformed data (pseudo anomalies). Therefore, we extract the features before the fc layer and then use features learned from normal training data to fit a multivariate Gaussian to get mean $\mu$ and covariance $\Sigma$.

\section{Experiments and results}

\begin{table}[t]
% \vspace{-0.3cm}
\centering
\caption{The statistics of three EEG datasets.}
\scalebox{0.95}{
\begin{tabular}{c|ccccc} 
\hline
Dataset  & ~Sampling rate~ & ~Channels~ & ~Patients~ & \begin{tabular}[c]{@{}c@{}}~Normal EEGs~\\(seconds)\end{tabular} & \begin{tabular}[c]{@{}c@{}}~Abnormal EEGs~\\ (seconds)\end{tabular}  \\ 
\hline
Internal & 1024            & 19         & 50         & 30008                                                            & 14402                                                                \\
CHB-MIT  & 256             & 18         & 21         & 70307                                                            & 11506                                                                \\
UPMC     & 400             & 16         & 4          & 9116                                                             & 1087                                                                 \\
\hline
\end{tabular}
}
	    \label{tab:datasets}
\end{table}

\noindent \textbf{Experimental settings:}
Three EEG datasets were used to evaluate the proposed approach, including the public Children’s Hospital Boston–Massachusetts Institute of Technology  Database (`CHB–MIT')~\cite{AliHossamShoeb2009ApplicationOM} and the UPenn and Mayo Clinic’s Seizure Detection Challenge  dataset (`UPMC')~\cite{AndriyTemko2015DetectionOS}, and an internal dataset from a national hospital (`Internal'). See Table~\ref{tab:datasets} for details. %While only the onset of epilepsy (i.e., seizure) is included as anomaly in the two public datasets, seven statuses (including seizure) of epilepsy were included in the internal dataset.%
While only the ictal stage of seizure is included as anomaly in UPMC, both ictal and various interictal epileptiform discharges (IEDs) were included in Internal and CHB-MIT. Particularly, the abnormal waveforms in Internal include triphasic waves, spike-and-slow-wave complexes, sharp-and-slow-wave complex, multiple spike-and-slow-wave complexes,  multiple sharp-and-slow-wave complex and ictal discharges.
%which therefore is likely more challenging for anomaly detection. 
2 out of 23 patients were removed from CHB–MIT due to irregular channel naming and electrode positioning. For normal EEG recordings lasting for more than one hour in CHB-MIT, only the first hour of normal EEG recordings were included to partly balance data size across patients. In UPMC, only dog data were used because of inconsistent recording locations across human patients.
For CHB–MIT and Internal, each original EEG recording was cut into short segments of fixed length (3 seconds in experiments). %; \textbf{results with 1 and 2 seconds included in Supplementary}. 
For UPMC, one-second short segments have been provided by the organizer. Each short segment was considered as one EEG data during model development and evaluation. Therefore the size of each EEG data is  [number of channels, sampling rate $\times$ segment duration].
%Due to different sampling rates and EEG channels among the three datasets, the number of sampling values along the time dimension is   
For each dataset, signal amplitude in EEG data was normalized to the range $[0, 1]$ based on the minimum and maximum signal values in the dataset.

On each dataset, while all abnormal EEG data were used for testing, normal EEG data were split into training and test parts in two ways. One way (Setting I, default choice) is to randomly choose the same number of normal EEGs as that of abnormal EEGs for testing and the remaining is for training, without considering patient identification of EEGs. The other (Setting II, subject level) is to split patients with the cross-validation strategy, such that all normal EEGs of one patient were used either for training or test at each round of cross validation. 

In training the 3-class classifier, for each batch of 64 normal EEGs, correspondingly 64 amplitude-abnormal EEGs and 64 frequency-abnormal (32 higher-frequency and 32 lower-frequency) EEGs were generated (see Section~\ref{SSL-Generation}). Adam optimizer with learning rate 0.0001 and weight decay coefﬁcient 0.00003 were adopted, and training was consistently observed convergent within 300 epochs. %In experiments,
$[\alpha_l, \alpha_h] = [2.0, 4.0]$, $[w_l, w_h] = [4, L]$, $[\omega_l, \omega_h] = [2, 4]$, and $[\omega'_l, \omega'_h] = [0.1, 0.5]$. The ranges of amplitude and frequency scalar factors were determined based on expert knowledge about potential changes of abnormal brain wave compared to normal signals. The area under ROC curve (AUC) %was used for performance measure, 
and its average and standard deviation over five runs (Setting I) or multiple rounds of validation (Setting II), the equal error rate (EER), and F1-score (at EER) were reported.  

\begin{figure}[t]
\centering
\includegraphics[width=0.6\linewidth,height=0.23\linewidth]{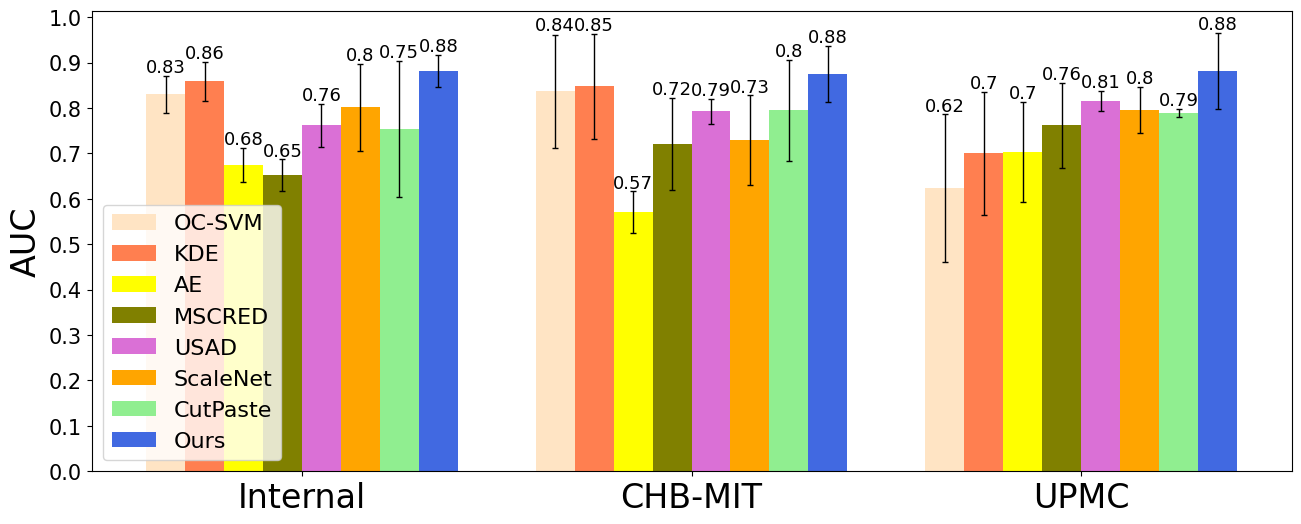}
\includegraphics[width=0.36\linewidth,height=0.23\linewidth]{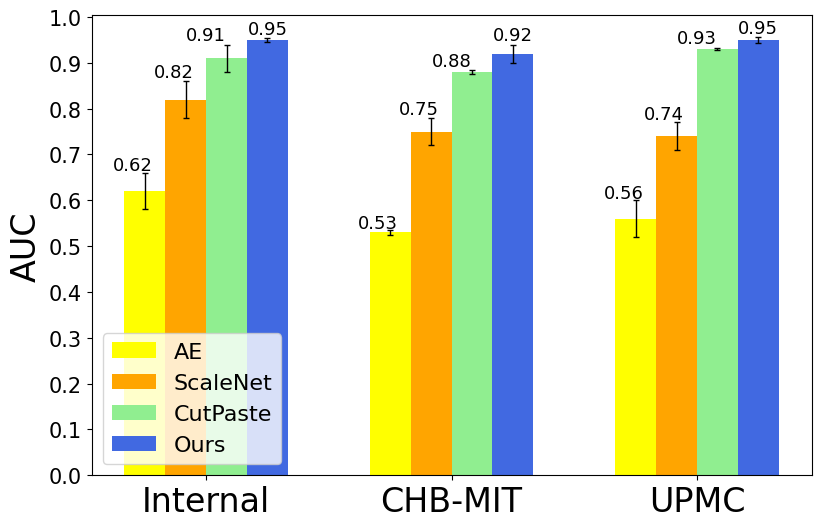}
\caption{Performance comparison 
%with cross-validation strategy 
at the subject level (i.e., Setting II) 
%for anomaly detection 
on each dataset (Left), and the ablation study of the feature extractor training (Right). AUC values were included in figure for each method. Vertical line: standard deviation.}
%\vspace{-0.5cm}
\label{fig:effective_subject}
\end{figure}

\noindent \textbf{Effectiveness evaluation:}
Our method was compared with  well-known anomaly detection methods including the one-class SVM (OC-SVM)~\cite{BernhardSchlkopf1999SupportVM}, the statistical kernel density estimation (KDE), and the autoencoder (AE)~\cite{ZhaominChen2018AutoencoderbasedNA}, the recently proposed methods Multi-Scale Convolutional Recurrent Encoder-Decoder (MSCRED)~\cite{zhang2019deep} and Unsupervised Anomaly Detection  (USAD)~\cite{audibert2020usad} for multivariate time series, and the recently proposed SSL methods for anomaly detection, including ScaleNet~\cite{JunjieXu2020AnomalyDO} and CutPaste~\cite{ChunLiangLi2021CutPasteSL}.
Note that ScaleNet~\cite{JunjieXu2020AnomalyDO} uses frequencies of normal EEGs at multiple scales to help detect abnormal EEGs, without considering any characteristics in abnormal EEGs.
%alters normal EEG only in terms of frequency, using classification probabilities to calculate anomaly scores, and it does not actually take into account expert knowledge of abnormal EEG. While our method constructs abnormal EEG in terms of magnitude and frequency, which aims to obtain a better feature extractor that can effectively extract discriminative features to help discriminate normal from abnormal EEGs.
Similar efforts were taken to tune relevant hyper-parameters for each method. In particular, to obtain feature vector input for OC-SVM and KDE, every row in each EEG data was reduced to a 64-dimensional vector by principal component analysis (PCA) based on all the row vectors of all normal EEGs in each dataset, and then all the dimension-reduced rows were concatenated as the feature representation of the EEG data.
For AE, the encoder consists of three convolutional layers and one fully connected (FC) layer, and symmetrically the decoder consists of one FC and three deconvolutional layers.
For CutPaste, each normal EEG in each training set is considered as a gray image of size $K \times L$ pixels, and the suggested hyper-parameters from the original study~\cite{ChunLiangLi2021CutPasteSL} were adopted for model training. For ScaleNet, the method was re-implemented with suggested hyper-parameters~\cite{JunjieXu2020AnomalyDO}. 
As Table~\ref{tab:effective} shows, on all three datasets, our method (last row) outperforms all the baselines by a large margin. Consistently, as Figure~\ref{fig:effective_subject} (Left) demonstrates, our method performed best as well at the subject level (i.e., Setting II), although the performance decreases a bit due to the more challenging setting. All these results
clearly confirm the effectiveness of our method for anomaly detection in EEGs. 
%*****(\textbf{may desribe certain observation in results, such as slightly worse performance on Internal dataset due to multiple types of anomalies})***********.

\begin{table}[!tbp]
 \caption{Performance comparison on three datasets with Setting I. %Each EEG data is 3 seconds long for  CHB-MIT and  Internal datasets. 
 Bold face indicates the best, and italic face for the second best. Standard deviations are in brackets.}
% \vspace{-0.3cm}
  \centering
  \resizebox{\textwidth}{!}{
%   \begin{tabular}{lcccccccccccc}
%     \toprule
%     % \small
%      \multirow{2}{*}{\textbf{Method}}&\multicolumn{3}{c}{\textbf{RSNA Setting-1}}&\multicolumn{3}{c}{\textbf{RSNA Setting-2}}&\multicolumn{3}{c}{\textbf{RSNA Setting-3}}&\multicolumn{3}{c}{\textbf{Pediatric}} \\
%      & EER$\downarrow$ & F1$\uparrow$ & AUC$\uparrow$ & EER$\downarrow$ & F1$\uparrow$ & AUC$\uparrow$ & EER$\downarrow$ & F1$\uparrow$ & AUC$\uparrow$ & EER$\downarrow$ & F1$\uparrow$ & AUC$\uparrow$ \\
%      \hline
%     AE & 0.36 & 0.64 & 0.68 & 0.40 & 0.60 & 0.63 & 0.38 & 0.62 & 0.65 & 0.41 & 0.65 & 0.64 \\
%     OC-SVM-1 & 0.41 & 0.59 & 0.63 & 0.45 & 0.54 & 0.57 & 0.42 & 0.58 & 0.60 & 0.38 & 0.67 & 0.67 \\
%     OC-SVM-2 & 0.31 & 0.69 & 0.74 & 0.40 & 0.60 & 0.64 & 0.46 & 0.64 & 0.69 & 0.39 & 0.66 & 0.68\\
%     f-AnoGAN & \textit{0.21} & \textit{0.79} & \textit{0.84} & \textit{0.31} & \textit{0.68} & \textit{0.73} & \textit{0.27} & \textit{0.73} & \textit{0.79} & \textit{0.33} & \textit{0.72} & \textit{0.71}\\
%     Ours & \textbf{0.18} & \textbf{0.81} & \textbf{0.89} & \textbf{0.28} & \textbf{0.72} & \textbf{0.78} & \textbf{0.22} & \textbf{0.77} & \textbf{0.83} & \textbf{0.29} & \textbf{0.75} & \textbf{0.78}\\
% 	\bottomrule
\begin{tabular}{c|ccclccclccc} 
\hline
\multirow{2}{*}{\textbf{Method}} &                     & \textbf{\textbf{Internal}} & \multicolumn{1}{l}{} &  &                     & \textbf{\textbf{CHB-MIT}} & \multicolumn{1}{l}{} &  & \multicolumn{3}{c}{\textbf{UPMC}}                                  \\ 
\cline{2-4}\cline{6-8}\cline{10-12}
                                 & EER$\downarrow$     & F1$\uparrow$               & AUC$\uparrow$        &  & EER$\downarrow$     & F1$\uparrow$              & AUC$\uparrow$        &  & EER$\downarrow$      & F1$\uparrow$        & AUC$\uparrow$         \\ 
\cline{1-4}\cline{6-8}\cline{10-12}
\multicolumn{1}{r|}{OC-SVM}      & 0.30(0.008)         & 0.71(0.008)                & 0.75(0.02)           &  & 0.33(0.001)         & 0.69(0.001)               & 0.73(0.003)          &  & 0.33(0.01)           & 0.51(0.02)          & 0.74(0.01)            \\
KDE                              & 0.24(0.001)         & 0.76(0.001)                & 0.87(0.001)          &  & 0.32(0.002)         & 0.69(0.002)               & 0.75(0.003)          &  & 0.24(0.003)          & 0.70(0.004)         & 0.83(0.003)           \\
AE                               & 0.35(0.01)          & 0.65(0.01)                 & 0.69(0.02)           &  & 0.46(0.007)         & 0.54(0.007)               & 0.56(0.01)           &  & 0.32(0.02)           & 0.61(0.02)          & 0.75(0.02)            \\
MSCRED                              & 0.37(0.02)          & 0.64(0.03)                 & 0.67(0.03)           &  & 0.34(0.03)         & 0.67(0.03)               & 0.72(0.03)           &  & 0.28(0.02)           & 0.70(0.02)          & 0.76(0.02)   \\
USAD                           & 0.25(0.02)          & 0.76(0.02)                 & 0.83(0.02)           &  & 0.30(0.02)         & 0.69(0.03)               & 0.79(0.02)           &  & 0.24(0.02)           & 0.75(0.02)          & 0.82(0.02)             \\
ScaleNet                         & \textit{0.18(0.01)} & \textit{0.82(0.01)}        & \textit{0.90(0.01)}  &  & 0.26(0.03)          & 0.73(0.03)                & 0.81(0.03)           &  & \textit{0.20(0.03)}  & \textit{0.75(0.03)} & \textit{0.89(0.03)}   \\
CutPaste                         & 0.26(0.02)          & 0.74(0.02)                 & 0.83(0.03)           &  & \textit{0.26(0.01)} & \textit{0.74(0.01)}       & \textit{0.83(0.01)}  &  & 0.21(0.01)           & 0.74(0.01)          & 0.87(0.006)           \\
Ours                             & \textbf{0.11(0.01)} & \textbf{0.89(0.01)}        & \textbf{0.95(0.004)} &  & \textbf{0.16(0.01)} & \textbf{0.84(0.02)}       & \textbf{0.92(0.02)}  &  & \textbf{0.13(0.007)} & \textbf{0.83(0.01)} & \textbf{0.95(0.006)}  \\
\hline
\end{tabular}
  }
  \label{tab:effective}
\end{table}

%\subsection{Ablation and sensitivity studies}
\noindent \textbf{Ablation studies:}
% ablation study on all three datasets
% ablation 1: other feature extractor training (contrastive learning, (PCA), AE, Scale) at 1st stage + various strategies at 2nd stage vs. our FE at 1st stage + corresponding strategies at 2nd stage
To specifically confirm the effectiveness of the proposed task-oriented SSL strategy in training a feature extractor for EEG anomaly detection, 
an ablation study is performed by replacing this SSL strategy with several other SSL strategies, including (1) training an autoencoder and then keeping the encoder part as the feature extractor, % AE structure...
(2) training a ScaleNet and then keeping its feature extractor part whose structure is the same as the proposed one, and (3) contrastive learning of the proposed two-branch feature extractor using the well-known SimCLR method~\cite{TingChen2020ASF}. As
%Table~\ref{tab:ablation_ssl} 
Figure~\ref{fig:effective_subject} (Right) shows, compared to these SSL strategies which do not consider any property of anomalies in EEGs, % during feature extractor training,
our SSL strategy performs clearly better on all three datasets.
Another ablation study was performed to specifically confirm the role of the proposed two-branch backbone and self-labeled abnormal EEGs during feature extractor training.  From Table~\ref{tab:ablation_backbone}, it is clear that the inclusion of the two types of simulated anomalies boosted the performance from AUC=0.71 to 0.93, and additional inclusion of the second branch (`Two-branch') and change of kernel size from $1 \times 3$ to $1 \times 7$ (`Larger kernel') further improved the performance.% (AUC=0.**).

%\textcolor{red}{In addition, to confirm that the proposed simple transformations are sufficient to help train an anomaly detector, one more experiment was performed by training the feature extractor with simulated abnormal data based on more complex transformations (i.e., combinations of simulated abnormal frequency & amplitude rather than ) rules could performed better than the more complex rules, an ablation study was performed in 'Internal' dataset and the CHB-MIT dataset by training feature extractor with more complex SSL rules based on combinations of simulated abnormal frequency and amplitude. It results in unsatisfactory 0.90(0.01) AUC in 'Internal' dataset and 0.79(0.28) AUC in the CHB-MIT dataset. Although simulated combinations of anomalies are more realistic, they may not cover all possible anomalies in real abnormal EEGs and so model training would cause overfitting to limited simulated complex anomalies. The proposed SSL rules, which improve the power of feature extractor in extracting discriminative features, outperform more complex SSL rules with AUC 0.95(0.004) in 'Internal' dataset and AUC 0.92(0.02) in the CHB-MIT dataset).}
% In addition, one may expect that ... While real abnormal EEGs are often complex and include combinations of abnormal frequency and amplitude signals, 
In addition, one more evaluation showed the proposed simple transformations (based on simulated anomalies in frequency and amplitude individually) performed better than the more complex transformations (based on combinations of simulated abnormal frequency and amplitude), with AUC 0.954 vs. 0.901 on dataset Internal, 0.924 vs. 0.792 on CHB-MIT, and 0.952 vs. 0.918 on UPMC. Although simulated combinations of anomalies could be more realistic, they may not cover all possible anomalies in real EEGs and so the feature extractor could be trained to extract features of only the limited simulated complex anomalies. In contrast, with simple transformations, the feature extractor is trained to extract features which are discriminative enough between normal EEGs and simulated abnormal EEGs based on only abnormal frequency or amplitude features, therefore
%i.e., the feature extractor trained with basic SLL rules is 
more powerful and effective in extracting discriminative features. % (because it can extract discriminative features based on either aspect of anomalies rather than combinations of two aspects).

\begin{table}[t]
\caption{The effect of simulated anomaly classes and the two-branch architecture on anomaly detection with ‘Internal’ dataset. %The first result (0.71) is from ScaleNet with ResNet34 backbone. %Consistent results were obtained on the other two datasets.
}
%\vspace{-0.3cm}
    \centering
    \resizebox{\textwidth}{!}{
    \setlength{\tabcolsep}{1.0mm}{
   \begin{tabular}{l|ccccccc}
   \hline
Amplitude-abnormal &             & \checkmark           &            & \checkmark          & \checkmark           & \checkmark           & \checkmark            \\
Frequency-abnormal &            &             & \checkmark          & \checkmark          & \checkmark           & \checkmark           & \checkmark            \\
Two-branch         &            &             &            &            & \checkmark           &             & \checkmark            \\
Larger kernel      &            &             &            &            &             & \checkmark           & \checkmark            \\ 
\hline
AUC                & 0.71(0.04) & 0.91(0.001) & 0.89(0.01) & 0.93(0.01) & 0.94(0.006) & 0.94(0.005) & 0.95(0.004)  \\
\hline
\end{tabular}
    }
}
    \label{tab:ablation_backbone}
\end{table}

\begin{figure}[t]
\centering
\includegraphics[width=0.23\textwidth]{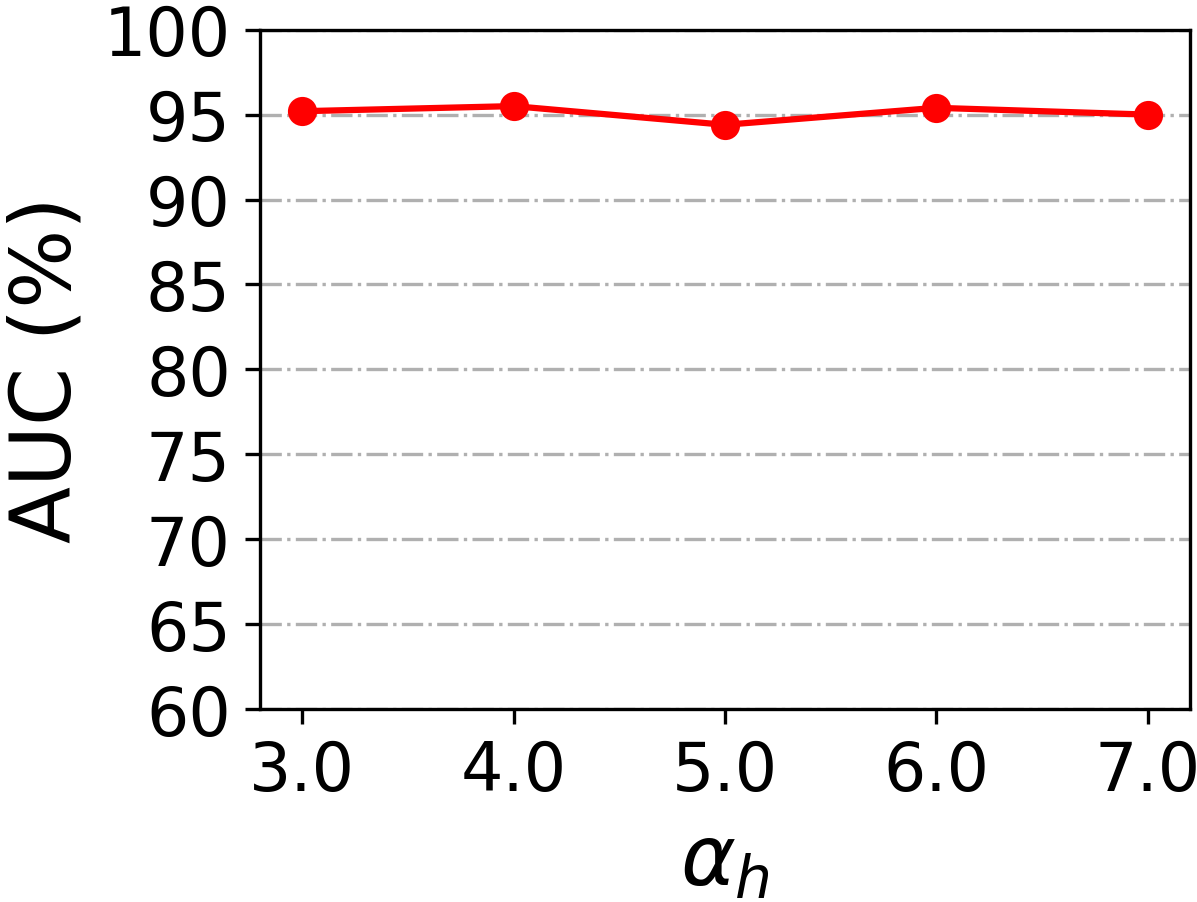}
\includegraphics[width=0.23\textwidth]{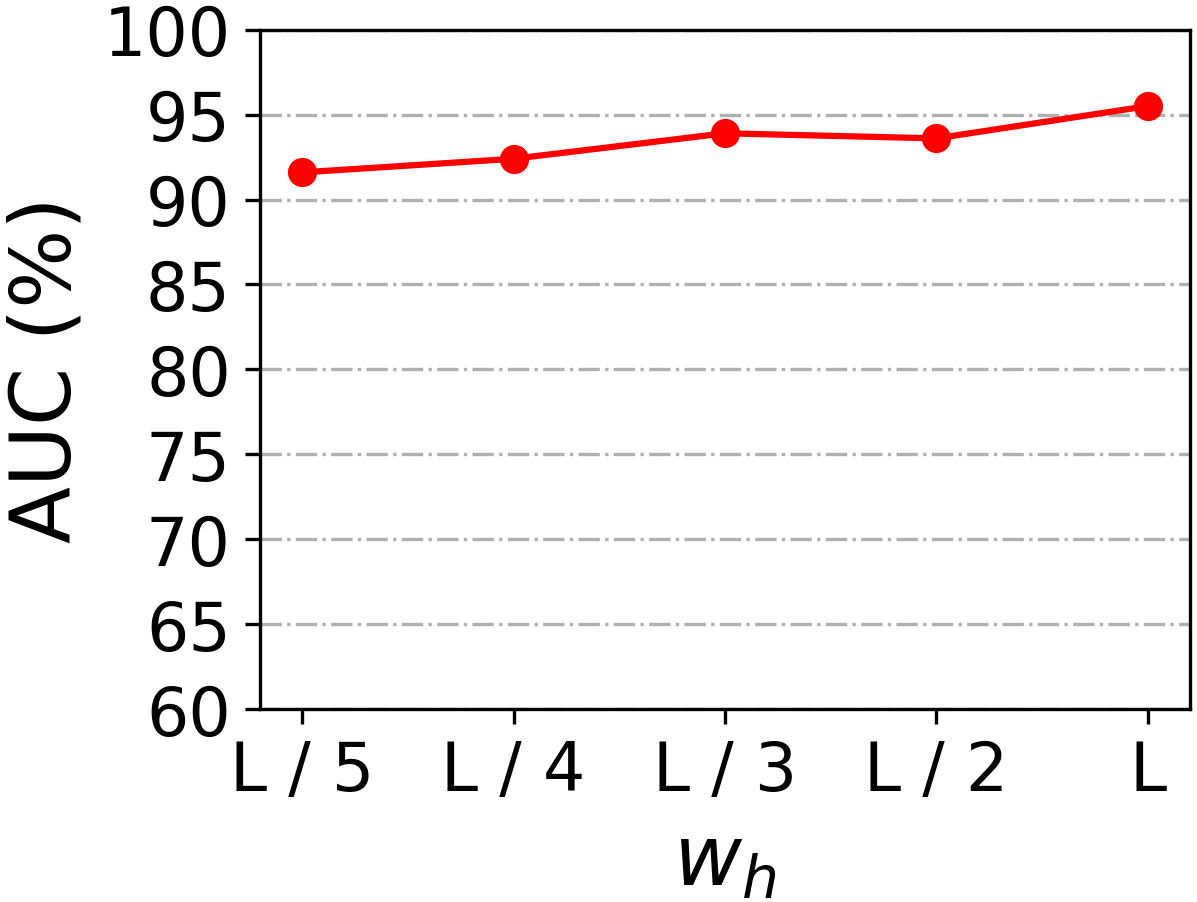}
\includegraphics[width=0.23\textwidth]{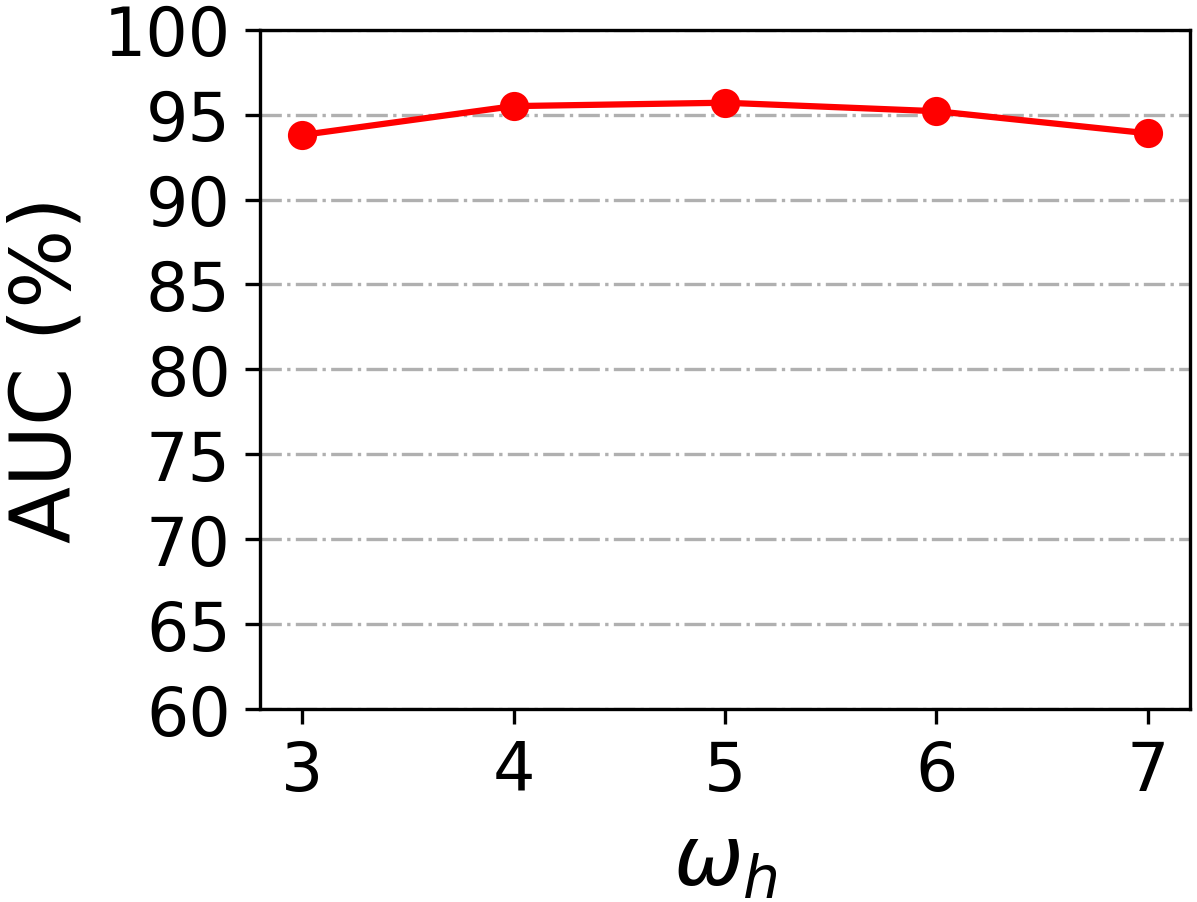}
\includegraphics[width=0.23\textwidth]{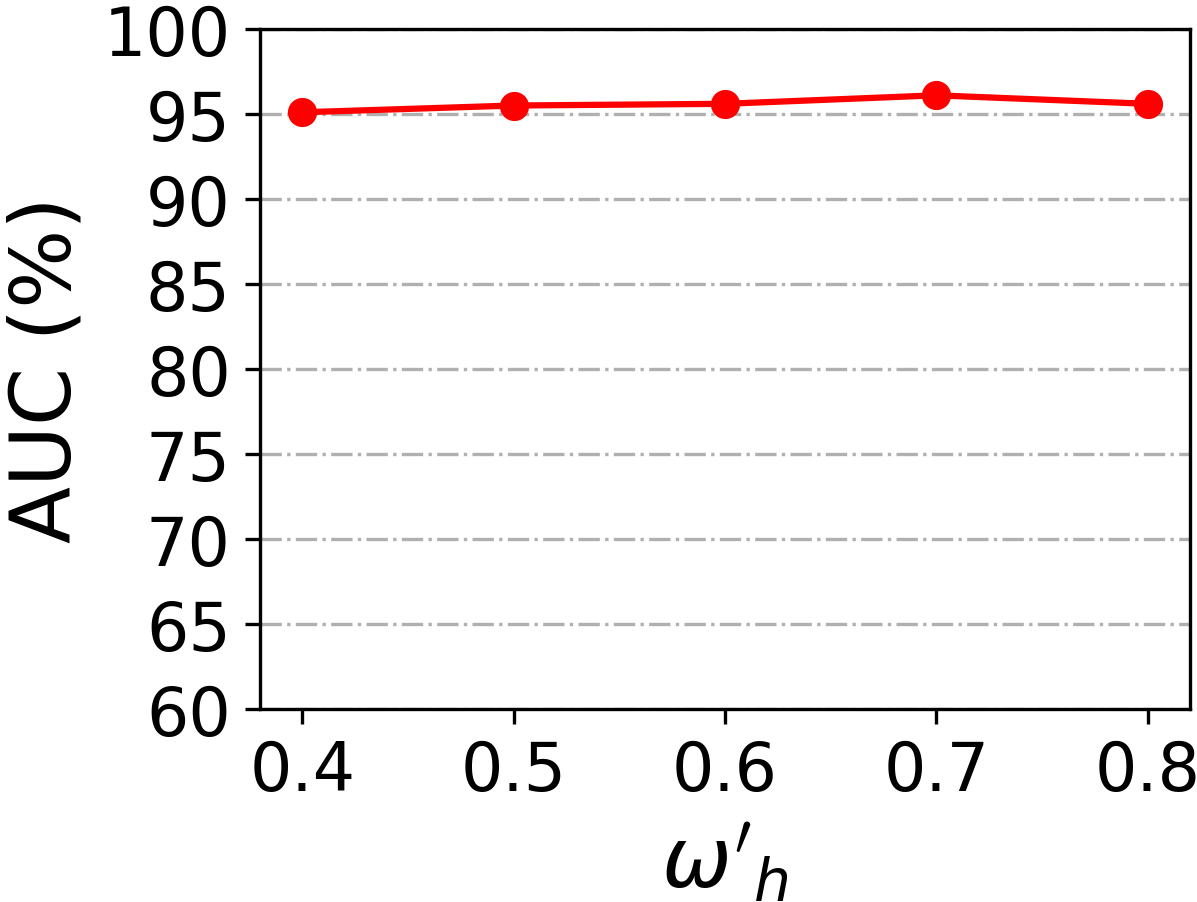}
% \begin{minipage}[]{0.50\linewidth}
% \centering
% %\subfloat[ $\lambda_a$]{\label{Fig:lambdaa}
% \includegraphics[height=0.7\textwidth,width=\textwidth]{alpha.png}
% %}

% \end{minipage} ~
% \begin{minipage}[]{0.50\linewidth}
% \centering
% %\subfloat[ $\lambda_a$]{\label{Fig:lambdaa}
% \includegraphics[height=0.7\textwidth,width=\textwidth]{w.png}
% %}

% \end{minipage} ~
% \begin{minipage}[]{0.50\linewidth}
% \centering
% %\subfloat[ $\lambda_a$]{\label{Fig:lambdaa}
% \includegraphics[height=0.7\textwidth,width=\textwidth]{omega.png}
% %}

% \end{minipage} ~
% \begin{minipage}[]{0.50\linewidth}
% \centering
% %\subfloat[$\lambda_u$]{\label{Fig:lambdau}
% \includegraphics[height=0.7\textwidth,width=\textwidth]{omega'.png}
% %}

% \end{minipage}
%\vspace{-0.3cm}
\caption{Sensitivity study of hyper-parameters. %Average and the last classification accuracy are shown over different hyper-parameter values on CIFAR100-B0 with 10 rounds. 
}
\label{fig:sen}
\end{figure}

% sensitivity study: hyper-param ranges - different ranges, kernel size 3-15 
\noindent \textbf{Sensitivity studies:}
The proposed task-oriented SSL strategy is largely insensitive to the hyper-parameters for generating simulated abnormal EEGs. For example, {as shown in Figure~\ref{fig:sen},} when respectively varying the range $[\alpha_l, \alpha_h]$ from $[2.0,3.0]$ to $[2.0,7.0]$, the range $[w_l, w_h]$ from $[4,L/5]$ to $[4,L]$, the range $[\omega_l, \omega_h]$ from $[2,3]$ to $[2,7]$, and the range $[\omega'_l, \omega'_h]$ from $[0.1,0.4]$ to $[0.1,0.8]$, the final anomaly performance changes in a relatively small range and all of them are clearly better than the baselilne methods.
%that of the feature extractor with the same two-branch backbone but trained using the ScaleNet strategy. 
These results support that the proposed task-oriented SSL strategy is quite stable in improving anomaly detection.

In addition, it is expected the proposed SSL strategy works stably even when injecting incompatible transformations. For example,
%We further study the robustness of our model by injecting incompatible rules. 
during training the feature extractor, when a proportion ($5\%$, $10\%$, $15\%$, $20\%$) of simulated abnormal EEGs were replaced by fake abnormal EEGs (each fake EEG randomly selected from real normal EEGs but used as abnormal), the AUC is respectively 0.927, % ($\pm 0.007$), 
0.916, % ($\pm 0.002$), 
0.898, % ($\pm 0.007$), 
and 0.883 %($\pm 0.003$) 
on Internal, lower than original 0.954 but still kept at high level.

% In extreme case, even when injecting incompatible rules, our SSL strategy should still work stably. E.g., during training feature extractor (Fig.2), when a proportion (5%, 10%, 15%, 20%) of simulated abnormal EEGs are replaced by fake abnormal EEGs (each fake EEG randomly from real normal EEGs but used as abnormal), the final performance (AUC) is respectively 0.938, 0.903, 0.899, and 0.891 on Internal, lower than reported AUC (0.954) but still kept at high level. 

\section{Conclusion}
In this study, we propose a two-stage framework for anomaly detection in EEGs  based on normal EEGs only. The proposed task-oriented self-supervised learning together with the two-branch feature extractor from the first stage was shown to be effective in helping improve the performance of the anomaly detector learned at the second stage. This suggests that although only normal data are available for anomaly detection in some scenarios, transformation of normal data with embedded key properties of anomalies may generate simulated abnormal data which can be used to greatly help develop a more effective anomaly detector. 

% \vspace{0.1cm}
\noindent \textbf{Acknowledgments}.
{This work is supported by NSFCs (No. 62071502, U1811461), the Guangdong Key Research and Development Program (No. 2020B1111190001), and the Meizhou Science and Technology Program (No. 2019A0102005).}

\bibliographystyle{splncs04} 
\bibliography{reference}

\end{document}